# Traction of Interlocking Spikes on a Granular Material

Volker Nannen[1] and Damian Bover[2]

[1]Sedewa, Finca Ecologica Son Duri, Vilafranca de Bonany, Spain.
Email: vnannen@gmail.com
[2]Sedewa, Finca Ecologica Son Duri, Vilafranca de Bonany, Spain.
Email: utopusproject@gmail.com

**ABSTRACT**

The interlock drive system generates traction by inserting narrow articulated spikes into the ground and by leveraging the soil's strength to resist horizontal draft forces. The system promises high tractive performance in low gravity environments where tires have little traction for lack of weight. At Earth and Space 2021 we reported the performance of such spikes on a silty clay loam, a cohesive soil. We found that in such soil, traction below a critical depth is provided by a zone of lateral soil failure. We also found that the articulation translates a horizontal draft force into a vertical penetration force strong enough to penetrate a narrow spike to a depth where the soil can sustain the draft force, in a self-regulating way. It is conceivable that a granular material like regolith or sand with little to no cohesive strength provides less vertical penetration resistance and less resistance to a horizontal draft force than a cohesive soil, which leads to the question of whether and how much tractive force an interlocking spike can generate on a granular material. Here we report on field trials that study different spike designs in dry and unsaturated moist sand. The results demonstrate that a loose granular material requires larger spikes than a cohesive soil, that these larger spikes penetrate dry and moist sand reliably, and that they promise good tractive efficiency. The trials indicate that on sand, a larger spike diameter can improve the pull/weight ratio without a loss of tractive performance.

**INTRODUCTION**

**Motivation.**  Tires generate traction on granular materials through friction with and between the upper particles of the material on which they move. On terrestrial soils, tires achieve an optimal tractive efficiency at a pull/weight ratio of about 0.4 (Zoz and Grisso 2003). In a lunar environment, Wilkinson and DeGennaro (2007) expect a lower pull/weight ratio of 0.21. Combined with low lunar gravity, this poses a serious challenge to the design of civil engineering vehicles like rippers, scrapers, and crawlers, as every kg of launch mass can develop a tractive force of only 0.35 N by friction on the Moon.

An alternative method to generate traction is to interlock spikes with the ground (Bover 2011; Nannen et al. 2016; 2017). As described by Nannen and Bover (2021), efficient and reliable ground penetration can be achieved if narrow spikes are attached to a lever arm that is attached to a hinge close to the ground. A backward force on such a spike drives it into the ground and a forward force pulls it out. Such interlocking spikes can then be incorporated into a push-pull vehicle where alternating frames push or pull tools like rippers or blades from the anchored spikes, see Figure 1.

Important design parameters for the spike are the diameter of the spike, the rake angle $\alpha$ between the soil surface and the spike, and the thrust angle $\gamma$ between the soil surface and a line from the spike tip to the hinge. $\gamma$ in turn depends on the spike radius r, which is the distance between the spike tip and the hinge, as well as the elevation of the hinge above the ground. See Table 1 for a list of symbols. The difference $\alpha - \gamma$ controls the vertical penetration force, which is maximized in the range $15° < \alpha - \gamma < 35°$. Since a horizontal draft force $F_D$ creates a lifting force $F_L = F_D \times \tan \gamma$ at the hinge, keeping the thrust angle small is important for vehicle stability. For a vehicle with spikes to pull twice its weight, $\gamma$ should not exceed 25°.

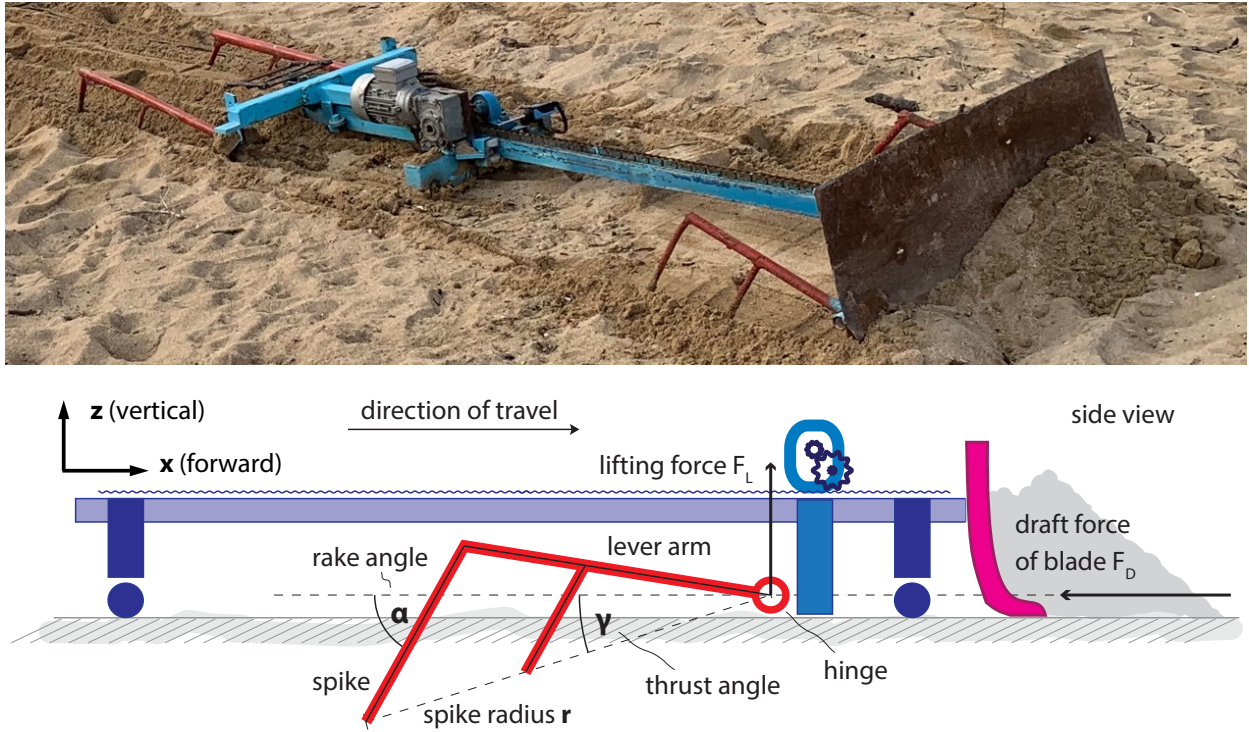

**Figure 1: Crawler with spikes for traction. Travel direction is from left to right. Top: demonstrator on beach sand. In an alternating motion pattern, the two spikes on the rear frame push the blade forward and the two spikes attached to the blade pull the vehicle forward. Bottom: simplified schematic drawing. Only one of the spikes that push the blade is shown.**

**Table 1: List of symbols.**

| | |
|---|---|
| $\alpha$ | rake angle, inclination from horizontal of the spike |
| $\gamma$ | thrust angle, inclination from horizontal of the line from hinge to spike tip |
| $F_D$ | draft force |
| $F_L$ | lifting force |
| $r$ | spike radius, the distance between spike tip and hinge |

The narrower the spike and the higher the rake angle, the higher the critical depth of a narrow spike, which is crucial for the generation of tractive force. The theory of critical depth states that there are two distinct modes of soil failure along a spike that is pulled laterally through the soil and that these two modes are vertically separated at a critical depth that depends on soil conditions, spike width, and rake angle (Zelenin 1950; Kostritsy 1956; O'Callaghan and McCullan 1965;

Published in Earth and Space 2022, doi.org/10.1061/9780784484470.009                                                     96

Hettiaratchi 1965; Godwin and Spoor 1977). Above the critical point, a narrow spike fragments and lifts a crescent of soil, which limits the available tractive force to the weight and friction of that crescent, see Figure 2. Below the critical point, the spike compresses the soil and pushes it forward and sideways, allowing for far greater tractive forces than would be possible with crescent failure.

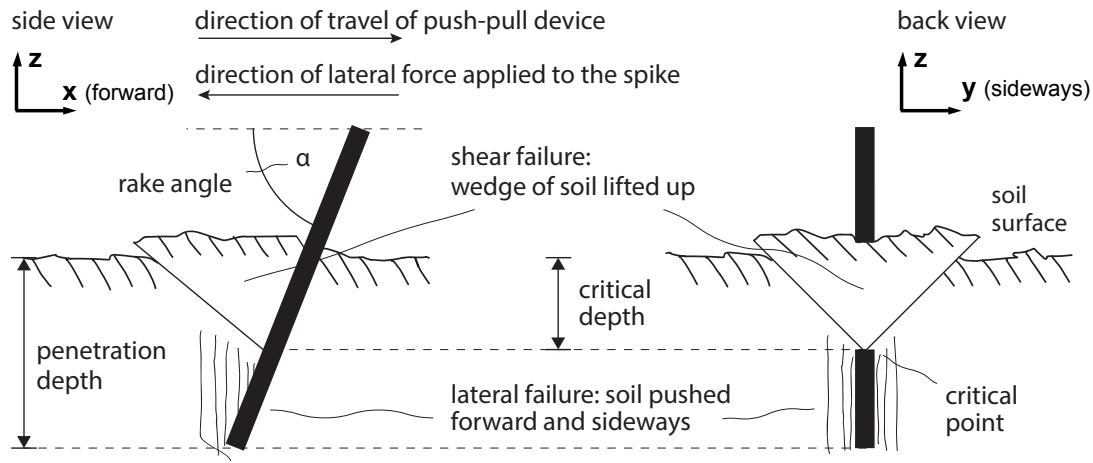

**Figure 2: different regimes of soil failure above and below the critical depth.**

Previous studies of the performance of spikes were done on cohesive agricultural soil, specifically on a silty clay loam with 6% gravel and stones by weight (Nannen ea., 2019, 2021). The present study wants to identify and evaluate spike designs that work reliably on a loose granular material.

**EXPERIMENTAL SETUP**

We tested one spike design with a small radius and several spike designs with a large spike radius. The large spike radius was tested in combination with four different spike diameters. The tests were conducted on dry and on unsaturated moist sand, as described below.

Both spike designs had an initial rake angle of 45° and positioned the hinge 7–10 cm above the ground. The smaller design had spikes made from 12 mm rebar and a spike radius of 58 cm. It was designed for a penetration depth of up to 15 cm, at which point the rake angle was 60° and the thrust angle was 25°. The larger design had the main spike made from a 21 mm steel rod with a spike radius from hinge to spike tip of 134 cm. It was designed for a maximum penetration depth of 50 cm, at which point the rake angle was 67.5° and the thrust angle was 26°. Both designs had a smaller spike about midway between the main spike and the hinge. Previous experience has shown that a second spike improves initial soil penetration and overall performance.

To test for the effect of spike diameter, pipes of varying diameters were pulled over the main spike to increase its thickness. In this way, four levels of spike thickness were tested with the large design: 21 mm, 34 mm, 42 mm, and 49 mm, corresponding to typical exterior diameters for commercial ½, 1, 1¼, and 1½ inch pipes. Note that the weight of the spike at its tip increased with spike diameter, which might have helped during initial ground penetration. For an overview of different design parameters, see Table 2.

Table 2: Tested Spike designs.

|  | Small radius | Large radius | | | |
| --- | --- | --- | --- | --- | --- |
| Spike diameter / thickness | 12 mm | 21 mm | 34 mm | 42 mm | 49 mm |
| Spike weight at tip | 0.7 kg | 2.9 kg | 3.7 kg | 3.8 kg | 4.1 kg |
| Spike radius | 58 cm | 1,34 cm | | | |
| Design depth | 15 cm | 50 cm | | | |
| Initial rake angle | 45° | 45° | | | |
| Rake angle at design depth | 60° | 67.5° | | | |
| Thrust angle at design depth | 25° | 26° | | | |

We conducted the field trials on fine-medium beach sand of bioclastic origin at the Bay of Alcúdia, Mallorca, Spain, at the end of September 2021. The trial location was about 1 m above sea level, 60 m from the waterline, and bordered an area of natural dunes. The sand had been compacted by foot traffic from beachgoers but was off-limit to vehicles. At the location, an upper layer of 5–8 cm of sand had been dried by sun and wind, while the lower layer consisted of unsaturated moist sand with 4% water content by weight. The unsaturated moist sand was visually homogeneous down to a depth of 1 m. The angle of repose of the dry sand was 30°, the same as for a sample of oven-dried sand. When cutting into the unsaturated moist sand with a spade, the sand would allow an angle of repose of 100° and more. When piled, the moist sand had an angle of repose of 47°. We measured densities of 1,720 kg/m$^3$ for oven-dry sand and 1,790 kg/m$^3$ for the unsaturated moist sand.

We prepared two sites for field trials, a moist site with the original unsaturated moist sand and a dry site consisting of a trench filled with dry surface sand. For the moist site, we scraped off the upper layer of dry loose sand. The exposed soil surface dried out but was not disturbed by foot traffic so only the upper 2–3 cm was affected. For the dry site, we dug a trench 1 m deep into the unsaturated moist soil. The trench had straight walls that tapered towards the bottom, such that the width of the trench was 50 cm at the top and 30 cm at 1 m depth. The length of the trench was 120 cm at the top and 100 cm at 1 m depth. We filled the trench with the upper 2-3 cm of loose dry sand from the surrounding surface area.

To control the draft force, we suspended a basket via a pulley from a large tripod, see Figure 3. A cable ran from the basket to the top pulley, down to a second pulley at the base, and then horizontally to the test vehicle and the spike. The two pulleys had a combined friction coefficient of 0.23. By manually adding iron weights to the basket, we controlled the draft on the test vehicle. A draft of 2 kN required a weight of 265 kg at the basket.

The vehicle that carried the spike consisted of a simple steel frame on rigid caster wheels and weighed 4 kg. Including ballast and accounting for the fact that the lever arm rested partly on the vehicle, the caster wheels carried a vehicle weight of 50 kg. To minimize friction, the caster wheels of the test vehicle rolled over a thin sheet of metal that rested on the sand. The metal sheet had a segment cut out from its center, to avoid interference with the movement of the spike. With a friction coefficient of 0.04, the rolling resistance of the vehicle does not exceed 20 N. We ignore this rolling resistance in the analysis.

During each trial, each time we added iron weights to the basket, we recorded the weight of the basket in kg, the horizontal motion of the vehicle at the hinge in mm, and the inclination of the spike arm in degrees with a precision of 0.1°. Parameters of interest like penetration depth and the amount of work needed to penetrate a spike into the soil were calculated through basic trigonometric functions and numerical integration. For post-analysis, we recorded each trial with a video camera and took photos during and after each trial.

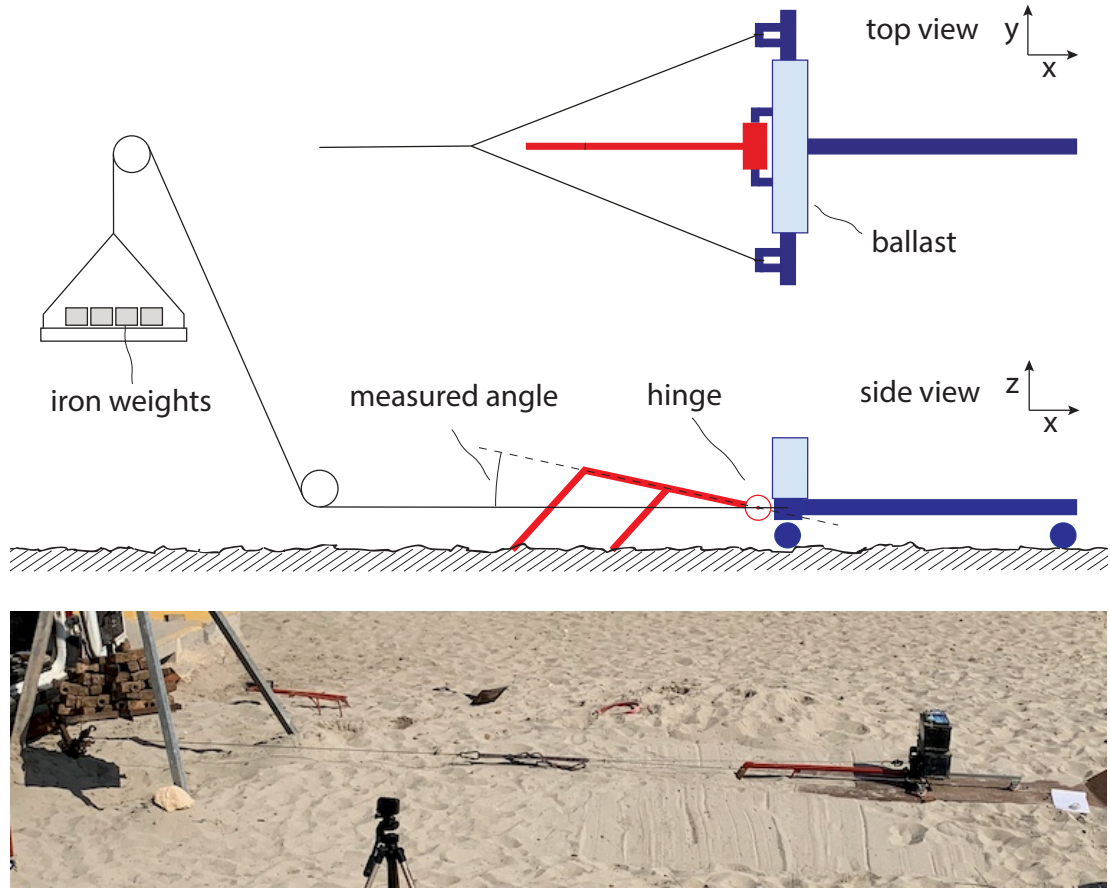

**Figure 3. Top: schematic view of the experimental setup. Bottom: photo of test location and setup. At the left are the iron weights in the basket. At the right is the test vehicle with a spike of a large radius.**

After conducting several unrecorded trials to assess the overall setup, we recorded the weight of the basket, the inclination of the spike, and the vehicle motion in 14 trials. Two trials assessed the performance of small spikes that perform well on cohesive agricultural soil. The remaining 12 trials assessed the performance of the larger spikes. Of those, one trial with a 34 mm spike was conducted with a blunt spike tip, the others with tapered tips. One trial was conducted without ballast on the vehicle, the others with a ballasted vehicle weight of 50 kg. In seven trials, the maximal draft force exceeded 2 kN and in one trial it almost reached 2 kN. The eight trials which almost reached or exceeded a draft force of 2 kN cover the four spike diameters, one each on dry and moist sand. They form the basis for the analysis and the graphs in the next section.

**EXPERIMENTAL RESULTS**

**Small spikes.** As an initial test, we recorded two trials on dry and moist sand with a spike design that works well on cohesive agricultural soil. These spikes were made from 12 mm rebar, had a spike radius of 58 cm, and were designed for a penetration depth of 15 cm. While the trench was 1 m deep for the other trials on dry sand, for this trial we prepared a trench of dry sand that was only 30 cm deep.

At a draft force of 1 kN, on moist sand, the penetration depth of the small spike was 18 cm, the thrust angle was 28°, and the vehicle slip was 22 cm. With the same draft force, on dry sand, the penetration depth was 37 cm—exceeding the depth of dry sand in the trench—the thrust angle was 52°, and the vehicle slip was 44 cm. We concluded that spikes of this size are insufficient for a tractive force of 1 kN on either dry or moist sand and conducted all remaining tests with the larger design.

**Tapered spike tips.** By pulling a pipe over the 21 mm rod that was shorter than the rod, we created a crudely tapered tip, in which the tip of the rod stuck out from the end of the pipe. Pulling a 38 cm long pipe over the 43 cm long rod created a difference in length of 5 cm. In all trials on dry sand, this combination of rod and pipe penetrated the sand without problems. On moist sand, on one occasion this combination slid over the sand without penetrating. We considered this a serious failure. Consequently, we shortened all pipes that we used to enlarge the diameter to 34 cm, for a difference in length between rod and pipe of 9 cm. This new design penetrated both the dry and the moist sand without a problem. We conclude that on moist sand the spike tip needs to be well tapered, while we could not make such a conclusion for dry sand.

**Ballast.** At a thrust angle $\gamma$, a horizontal draft force $F_D$ on the hinge creates a vertical lifting force $F_L$ at the hinge of $F_L = F_D \times \tan \gamma$. To protect the vehicle from flipping over, ballast needs to be added to counter this lift. With ballast, the weight of the vehicle was 50 kg. To test whether this amount of ballast was indeed necessary, we conducted one trial with a 21 mm spike on dry sand without ballast, such that only the vehicle frame and the lever arm countered the lift created at the hinge. The vehicle with a lever arm weighed 5 kg. At a penetration depth of 15 cm, the draft was 0.3 kN, the thrust angle was 9.5°, and the vehicle should in theory have been lifted. This did not happen. As we increased the draft, the vehicle was finally lifted off the caster wheels at a draft of 0.73 kN, a penetration depth of 34 cm, and a thrust angle of 18°. At this point, the lift should have been 240 N, enough to lift five times the vehicle's weight. During other trials where total vehicle weight was 50 kg—a gravitational Force of 490 N—we noted that the spike would at times enter deep enough to create a lift of 600 N or more without lifting the vehicle off its wheels. This suggests that unlike in cohesive soils, in sand the draft force is not centered at the spike tip but further up the spike, resulting in a significantly smaller thrust angle. As this might further improve the pull/weight ratio of the vehicle, it needs further investigation.

**Subsurface landslides.** During the trials, it became quickly apparent that the motion of the spike through the soil is not a smooth function of the increasing draft force. The draft could be increased by up to 450 N at a time without much effect on spike motion, at which point a further increase by 25 N would cause the device to be rapidly pulled forward by several centimeters and would cause the spike to rapidly penetrate several centimeters deeper into the sand. Apparently, stress builds up below the surface until the soil gives way rapidly in a subsurface landslide. This agrees well with the literature on landslides. The distribution over intervals between landslides will likely follow a power-law distribution (Dorogovtsev and Mendes, 1999) and needs further investigation.

The irregular sliding of the spike through the sand makes for graphs that are difficult to interpret. Figure 4 shows the penetration depth of the four spike diameters in dry and moist sand as a function of increasing draft force. As the internal stresses are released at different moments along the trajectory, the lines cross each other repeatedly, making it hard to decide whether spike diameter has any effect on penetration depth. For further analysis, we assume that the measurements of interest are those that were taken after each internal landslide because we assume that the forces have reached equilibrium after a landslide and because from an engineering point of view those are the maximum values that need to be accounted for—the maximal penetration depth, the maximal work expenditure and so on. Under this assumption, we can ignore measurements that were not recorded directly after a landslide and replace them by interpolating between the measurements taken directly after each landslide. This simplifies the graphs, as can be seen in Figure 5. The lines still mingle but show that spike diameter is not likely to affect initial penetration, but might reduce final penetration depth, which is important for the pull/weight ratio.

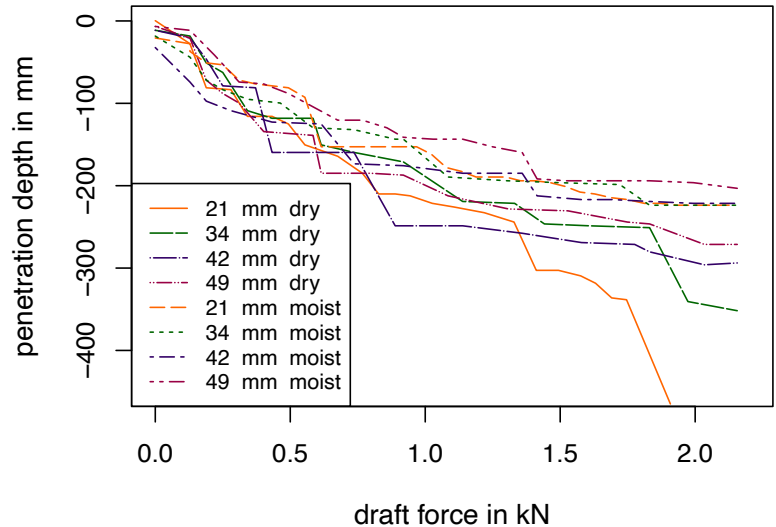

**Figure 4: Penetration depth for spikes of different diameters in moist and dry sand, calculated from the measured inclination of the lever arm.**

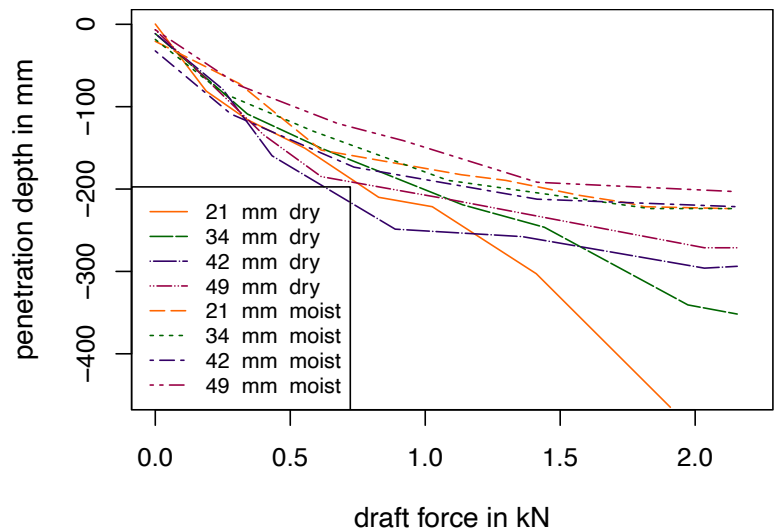

**Figure 5: Penetration depth for spikes of different diameters in moist and dry sand, calculated from the measured inclination of the lever arm after internal landslides, and interpolated between those measurements.**

**Efficiency and reliability.** Figure 6 shows the motion of the spike tip in the sand as calculated from the horizontal motion of the vehicle and the angular motion of the lever arm. This graph is based on all the original measurements, landslide or not. Despite the landslides, it shows that the motion of the spike through the soil is rather linear. It also shows that the angle of penetration is steeper for dry sand than for moist sand and that it decreases with spike diameter.

The energy required to penetrate the spike into the soil can be calculated by integrating over the product of the increments in draft force with the increment in horizontal spike motion. Figure 7 shows that this work is somewhat linear in the draft force, though there is some noise in the data. Note that the integrals were calculated from the original data before removing and interpolating data points between landslides.

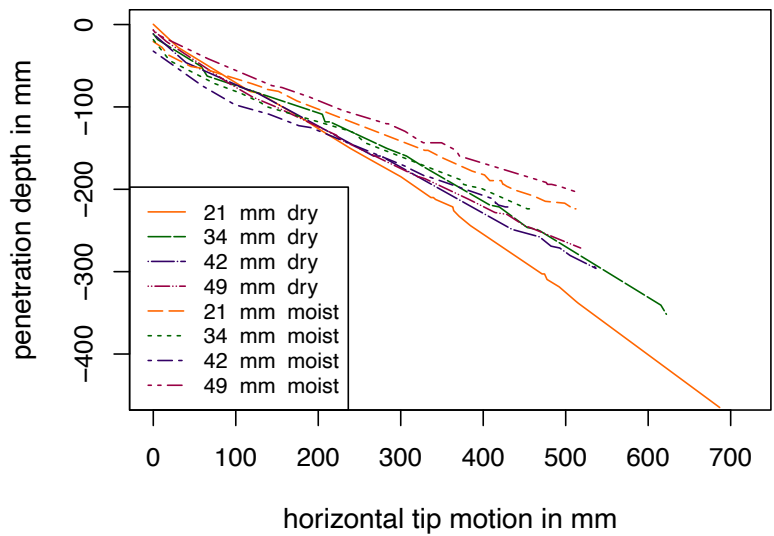

**Figure 6: The penetration depth at the tip is an almost linear function of horizontal motion at the tip.**

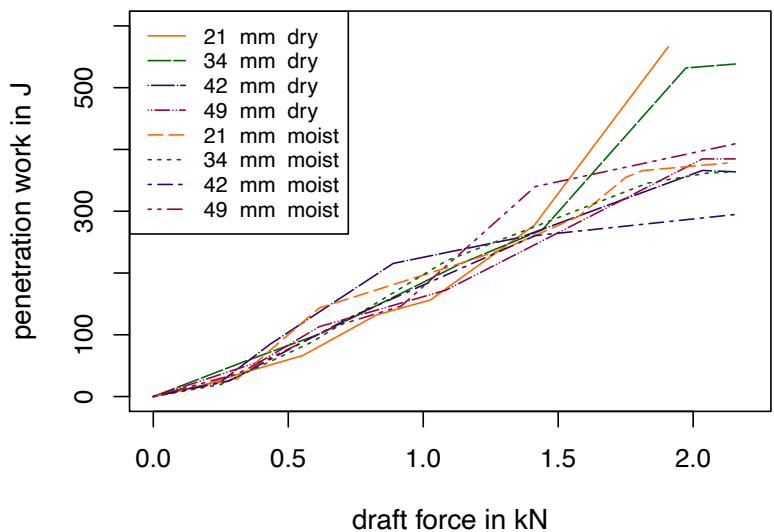

**Figure 7: The amount of work required to penetrate the spike into the ground so that it can withstand a given draft force is almost linear in the draft force.**

Tractive efficiency depends on penetration work: if it takes 350 J to penetrate a spike so that it can withstand a draft force of 2 kN, and then the blade of a crawler is pushed for 2 m from the spike with a force of 2 kN, the tractive efficiency at the spike is 4 kN / 4.35 kN = 92%, though overall vehicle efficiency will be less. If the required penetration work is linear in the draft force, the tractive efficiency of the spike is independent of the draft force. Figure 7, though noisy, does not suggest that spike diameter or soil moisture content have a significant effect on tractive efficiency.

**Vehicle stability.** A horizontal pull $F_D$ at the hinge of an interlocking spike creates a vertical lift $F_L$ at the hinge that is equal to $F_L = F_D \times \tan \gamma$, with $\gamma$ the thrust angle. To keep the vehicle stable, this lift needs to be countered with ballast and defines its pull/weight ratio. Figure 8 shows the thrust angle for a given draft force. Figure 9 shows the resulting lift. The graphs clearly show that lift is lower on moist sand and that there is a possibility that lift decreases with spike diameter for larger draft forces. This calls for further investigation.

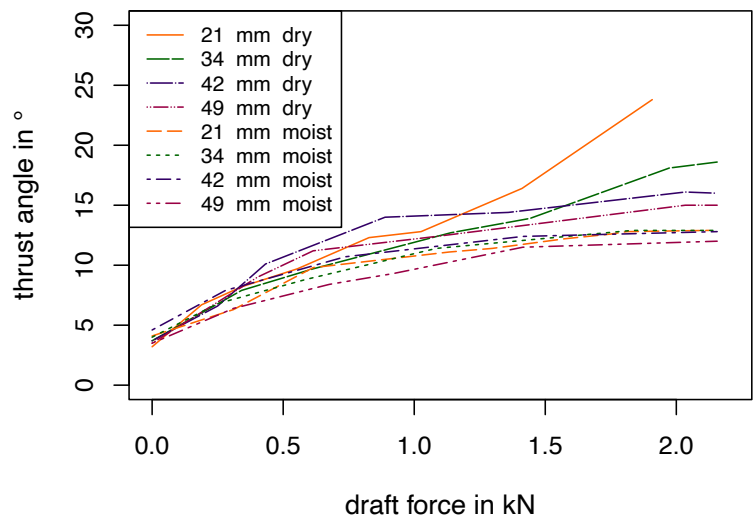

Figure 8: Thrust angle as a function of draft force $F_D$.

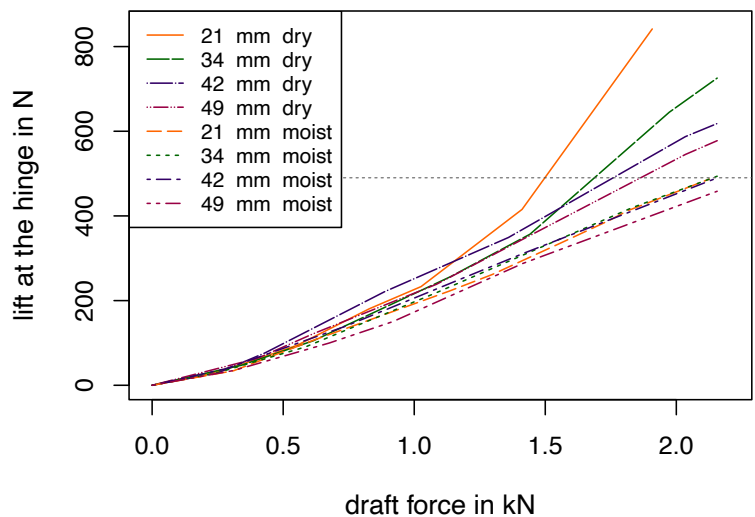

Figure 9: Calculated lifting force $F_L$ at the hinge in N. The dashed grey line shows when the calculated lifting force exceeds the vehicle weight.

## DISCUSSION AND CONCLUSION

To investigate the ability of interlocking spikes to generate traction on granular materials, we tested different spike designs on dry and unsaturated moist beach sand at a draft force of up to 2 kN. Once a spike of the correct dimension had been selected and the spike tip was sufficiently tapered, the spike penetrated dry and moist sand reliably and efficiently. Tractive efficiency seems to be independent of moisture content and spike diameter.

If the tractive force is the result of only crescent failure, the equilibrium forces can be calculated from the weight of the crescent and its friction along the shear plane. The angle of the shear plane is unknown. Numerical analysis shows that for the measured angles of repose, the horizontal pressure of the crescent against the spike is maximized at a shear plane angle of 45° for dry sand and 60° for moist sand. For dry sand, at a 45° shear plane angle and at the penetration depth we measured for a draft force of 2 kN, we calculate that a crescent of soil can press horizontally against the spike with a maximal force of 250 N for the 21 mm spike and of 40 N for the 49 mm spike. For moist sand, at a 60° shear plane angle, and at the penetration depth we measured for a draft force of 2 kN, we calculate that the maximal horizontal force of the crescent is 6 N for the 21 mm spike and 4 N for the 49 mm spike. We conclude that crescent failure cannot account for the observed tractive force of 2 kN. The tractive force is likely due to lateral soil failure as explained by the theory of critical depth.

Since the literature on critical depth clearly states that the critical depth decreases with spike thickness, we were concerned that spikes with larger diameters need to penetrate deeper to sustain the same draft force. This is not what we observed. Rather, our field trials indicate that larger diameter spikes need to penetrate less deep for the same tractive force. We also observed significantly less lift at the hinge than what can be calculated from the thrust angle, i.e., from the angle between a line from the hinge to the tip of the spike and the horizontal. This decrease in lift might be partly explained by the resistive force of loose sand against the rotational moment of the spike. A more likely explanation is, that in sand the draft force is not centered at the tip of the spike, as it seems to be in cohesive agricultural soil, but at a point higher up on the spike, resulting in a more favorable pull/weight ratio. Whether this is also true for lunar regolith needs to be established experimentally.